\documentclass[10pt]{extarticle}
\usepackage[a4paper, total={6in, 9in}]{geometry}
\usepackage[utf8]{inputenc}
\usepackage{natbib}

\usepackage{amsmath,amsfonts}
\usepackage{algorithmic}
\usepackage{algorithm}
\usepackage{array}
\usepackage{textcomp}
\usepackage{stfloats}
\usepackage{url}
\usepackage{verbatim}
\usepackage{graphicx}
\usepackage{cite}
\usepackage{caption}
\usepackage{subcaption}

\usepackage[utf8]{inputenc}
\usepackage[inline]{enumitem}
\usepackage[table]{xcolor}

\usepackage{multirow}
\usepackage{booktabs}
\usepackage{amssymb}

\usepackage{xspace}
\newcommand{\eg}{\textit{e.g.}\@\xspace}
\newcommand{\ie}{\textit{i.e.}\@\xspace}
\newcommand{\iid}{i.i.d.\@\xspace}

\newcommand{\R}{\mathbb{R}}
\newcommand{\E}{\mathbb{E}}

\newcommand{\pr}{\mathrm{p}}
\newcommand{\vect}[1]{\boldsymbol{\mathbf{#1}}}
\newcommand{\eqdef}{\doteq}

\newcommand{\gx}{\textsc{x}}
\newcommand{\gh}{\textsc{h}}
\newcommand{\gy}{\textsc{y}}
\newcommand{\vx}{\vect{x}}
\newcommand{\vh}{\vect{h}}
\newcommand{\vy}{\vect{y}}
\newcommand{\Vx}{\mathbb{V}_{\gx}}
\newcommand{\Vh}{\mathbb{V}_{\gh}}
\newcommand{\Vy}{\mathbb{V}_{\gy}}

\newcommand{\xth}{{h^x}}
\newcommand{\vxth}{\vh^{x}}

\newcommand{\vhth}{\widetilde{\vh}}

\title{\bfseries Graph State-Space Models\\and Latent Relational Inference}
\date{}

\usepackage{authblk}
\author[1]{Daniele Zambon\thanks{Corresponding author, \texttt{daniele.zambon@usi.ch} .}}
\author[2,1]{Andrea Cini}
\author[1,3]{Cesare Alippi}
\affil[1]{\small Universit\`a della Svizzera italiana, IDSIA, Switzerland.}
\affil[2]{\small EPFL, IMOS Lab, Switzerland.}
\affil[3]{\small Politecnico di Milano, Italy.}

\begin{document}

\maketitle

\begin{abstract}
State-space models effectively model multivariate time series by updating over time a representation of the system state from which predictions are made. 
The state representation is usually a vector without any explicit structure. Relational inductive biases, e.g., associated with dependencies among input signals and state representations, are not explicitly exploited during processing, leaving unattended opportunities for effective modeling. 
The manuscript aims to fill this gap by matching state-space modeling and spatio-temporal data where the relational information, say the functional graph capturing latent dependencies, is learned directly from time series.  
In particular, we propose \emph{Graph State-Space Models}, a novel probabilistic framework that jointly learns state-space dynamics and latent relational structures end-to-end on downstream tasks.
The proposed framework generalizes several state-of-the-art methods and, as we show, is effective in extracting meaningful latent relational structures and obtaining accurate forecasts.
\end{abstract}

\section{Introduction}\label{sec:introduction}

State space models have a long history in systems theory and signal processing, serving as a unifying framework for describing time-invariant dynamical processes through compact latent representations \citep{durbin2012time,box2015time}. Decades of research have shown that modeling a system in terms of its state and transition dynamics provides a principled approach to characterizing finite-memory data-generating processes, particularly those governed by differential equations. In their traditional discrete linear form, such models are typically written as  
\begin{equation}\label{eq:ssm-linear}
\begin{cases}
\vh_t = A\,\vh_{t-1} + B\,\vx_t + \eta_t \\
\vy_t = C\,\vh_t + \nu_t,
\end{cases}
\end{equation}
where $\vh_t\in\mathbb R^{d_h}$ denotes the latent state, $\vx_t\in\mathbb R^{d_x}$ external inputs, $\vy_t\in\mathbb R^{d_y}$ observed outputs, and $\eta_t,\nu_t$ uncertainty modeled as Gaussian processes. This classical formalization laid the foundation for modern sequence modeling, providing tools for analyzing stability, controllability, and identifiability properties \citep{sontag2013mathematical,box2015time}. 

Beyond linear formulations, substantial research has explored nonlinear state-space models and non-Gaussian processes, motivated by applications where assumptions of~\eqref{eq:ssm-linear} are inadequate for complex system dynamics \citep{kantz2003nonlinear,durbin2012time}.   
A probabilistic form for the nonlinear system
\begin{equation}\label{eq:ssm-prob}
\begin{cases}
\vh_t \sim \pr(\vh \mid \vh_{t-1}, \vx_t) \\
\vy_t \sim \pr(\vy \mid \vh_t)
\end{cases}
\end{equation}
supports richer temporal evolutions and output distributions, albeit often at the expense of computational tractability and theoretical guarantees. 
Nonlinear formulations significantly broaden the effectiveness of state-space approaches and have influenced a wide range of modern deep sequence architectures across domains, including control, dynamical systems identification, and machine learning---see, \eg, \citet{hochreiter1997long,chen2017maximum,brunton2016discovering,brunton2021modern, gu2022efficiently}.

In parallel, consolidating evidence demonstrates that incorporating relational or structural priors in predictive models provides effective inductive biases for several application domains and tasks, such as multistep forecasting \citep{seo2018structured,li2018diffusion} and missing-value imputation \citep{cini2021filling,marisca2022learning,chen2022adaptive}. Developments build on major advances in deep learning and signal processing for graph-structured data---see \citep{bronstein2017geometric,stankovic2019understanding,bacciu2020gentle,bronstein2021geometric} for overviews---and have inspired state-space architectures that jointly handle temporal and spatial dependencies \citep{behrouz2024graph,eliasof2025grama}. While relational information arises naturally from domain knowledge in many applications, in others the underlying graph is not observed, and the structural dependencies must be inferred directly from data \citep{kipf2018neural,kazi2022differentiable,cini2023sparse}.

Within this line of work, there remains significant room to enrich state-space models by leveraging learned relational representations. 
In particular, many existing approaches use state representations tied to the input and output dimensionalities. 
Indeed, even graph-based methods mostly rely on a one-to-one mapping between input time series and nodes in the learned representation. 
This implies that, in these methods, inferred relationships are limited to representing dependencies among the observed time series rather than relationships among latent factors. In other words, existing methods have not addressed the learning graph state-space models where the state of the system is represented by a latent attributed graph disjoint from input and output dimensions. 
This void leaves open opportunities to explicitly account for the uncertainty and temporal evolution of latent relational structures, as well as to leverage more flexible graph representations at different stages of the processing pipeline.

\begin{figure}
    \centering
    \includegraphics[width=.6\textwidth]{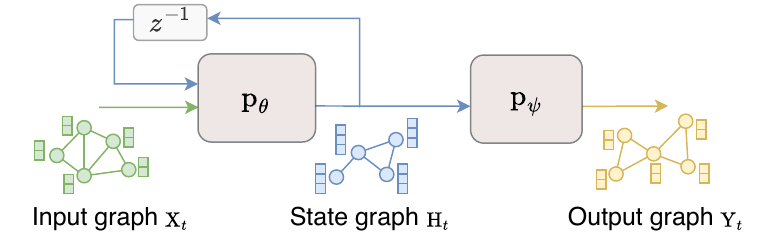}
    \caption{High-level representation of the predictive family of graph state-space models, following the structure of \eqref{eq:ssm-prob}. 
    State graph $\gh_t$ in blue is the processing outcome of input graph $\gx_t$ in green and previous state $\gh_{t-1}=z^{-1}(\gh_t)$; $z^{-1}$ being the lag (backshift) operator. Output graph $\gy_t$ in yellow follows from state graph $\gh_t$. Graphs are attributed with node-level signals, and they can present different topologies and number of nodes.}
    \label{fig:gssm_bare}
\end{figure}

In this paper, we introduce a generalized state-space formulation designed for spatio-temporal data modeling for settings where inputs, latent states, and outputs can each be characterized by a different graph structure. 
In contrast to traditional approaches, which rely on vector representations and fixed or shared relational structure, our formulation allows for much higher flexibility. 
This flexibility enables the model to learn unknown dependencies among both observed and latent variables directly from data. 
A schematic overview of the proposed formulation is shown in Figure~\ref{fig:gssm_bare}.
The main novel contributions of the paper can be summarized as follows:
\begin{itemize}
    \item We present a general probabilistic state-space formulation for spatio-temporal data modeling in which input, output, and state representations are attributed graphs with a possibly time-varying topology and node sets (Section~\ref{sec:model-formulation}).
    \item We provide an end-to-end learning procedure that accounts for the stochastic nature of state and output graphs. Latent graph states are directly learned from data via gradient-based optimization of parametric graph distributions (Section~\ref{sec:learning-strategy}).
    \item We show that several state-of-the-art models, as well as traditional approaches, used in different application domains arise as special cases of our formulation, providing a unifying perspective on existing methods (Section~\ref{sec:existing-instances}).
\end{itemize}
Empirical results on both synthetic and real-world datasets demonstrate that the proposed framework achieves competitive or better forecasting accuracy, while offering additional insights in terms of latent state dynamics and relational structure (Section~\ref{sec:experiments}). 

Overall, this methodological work advances state-space and spatio-temporal modeling by introducing a novel approach to learn latent relational representations from data and systematically incorporating them at all stages of the model pipeline. The {graph state-space} formulation introduced here offers a general framework to characterize state-space models operating on graph-structured dynamic data.

\section{Related Works}

The well-established state-space approach for modeling dynamical systems and time series finds applications across signal processing, statistics, and machine learning \citep{durbin2012time,sontag2013mathematical,haykin2001kalman}.  
Similar ideas have been revisited within deep learning by parametrizing transition and observation models with neural networks, yielding flexible nonlinear and probabilistic sequence models \citep{fraccaro2016sequential,krishnan2017structured,chen2018neural}. In parallel, renewed interest in linear state-space models has emerged as a means to improve different aspects, such as long-range dependency modeling, numerical stability, and computational efficiency in deep architectures \citep{rangapuram2018deep,gu2020hippo,gu2022efficiently,smith2023simplified,gu2024mamba}, with some approaches also addressing graph-structured data \citep{tang2023modeling,behrouz2024graph,li2024state,eliasof2025grama, ceni2025message, ding2025dygmamba}.

Large multivariate time series often consist of a collection of related time series, each of which is associated, for instance, with a specific geographical location or a node in a sensor network. 
In this setting, time series are usually correlated due to spatial proximity or shared dependencies, and graph deep learning has therefore emerged as a natural way to capture and exploit such relational structure as an inductive bias \citep{cini2025graph}. Models belonging to this framework, commonly referred to as spatio-temporal graph neural networks (STGNNs), condition predictions on observations from related nodes. This approach, pioneered in traffic forecasting \citep{li2018diffusion,yu2018spatiotemporal}, has achieved strong performance across a variety of tasks and applications \citep{fritz2022combining,marisca2022learning,iskandaryan2023graph,cini2023scalable}. Reviews of these architectures are available in the literature \citep{jin2024survey,chen2024graph}, and related approaches have also been developed from a signal processing perspective \citep{grassi2017time,stankovic2020data,leus2023graph}.

Most graph-based approaches rely on propagating information over a given graph that encodes known relationships or is derived from heuristics~(e.g., based on some similarity score). However, such relational information may be missing, inaccurate, or irrelevant for the task at hand. 
This has motivated approaches that learn graph structure directly from data and end-to-end alongside the predictive model.
Several methods model graphs as matrices whose entries parameterize the presence or strength of relations \citep{wu2019graph,bai2020adaptive,defelice2024graphbased} and are often combined with sparsification strategies \citep{wu2020connecting,deng2021graph,zhang2022graphguided}. Another line of work adopts a probabilistic perspective, modeling graphs as realizations of discrete distributions and enabling the incorporation of priors \citep{kipf2018neural,elinas2020variational,kazi2022differentiable,ahmed2022simple,cini2023sparse}. 
A few works have also addressed the learning of dynamic relations and hierarchical structure \citep{deng2021graph,marisca2024graph,hansen2025on}.

The approach adopted in this paper follows this direction of learning relational structures jointly with the remaining model parameters. 
However, in contrast to these works, the goal of this paper is not to propose a new forecasting architecture, but to introduce graph state-space models---a general probabilistic framework that unifies state-space modeling of temporal dynamics and relational learning. In particular, Section~\ref{sec:existing-instances} shows how existing approaches can be interpreted through the lens of graph state-space models.

\section{Graph State-Space Models}
\label{sec:model-formulation}

This section derives the proposed state-space model that generalizes vector-based formulations to the case where inputs, outputs, and states are graphs. As we will show, the generalization is not trivial, as it requires dealing with variable topologies over time and disjoint node sets, while keeping the processing localized and sparse where possible. Vector state representations can be seen as a special case in which no topological information is given, and nodes in the state representation collapse into a single entity.

\subsection{Model Formulation}
We extend the vector-based formalization of \eqref{eq:ssm-prob} to the graph case as
\begin{equation}\label{eq:gssm-system}
\begin{cases}
\gh_t \sim \pr(\gh \mid \gh_{t-1}, \gx_t) \\
\gy_t \sim \pr(\gy \mid \gh_t),
\end{cases}
\end{equation}
with initial state $\gh_0$ sampled from a prior distribution $\pr(\gh)$.
Inputs $\gx_t$, states $\gh_t$, and outputs $\gy_t$ for all time steps $t$ are all assumed graphs; such graphs belong, respectively, to spaces $\mathcal X,\mathcal H$, and $\mathcal Y$ of attributed graphs with a finite number of nodes and a (possibly) different and dynamic topology. 
A graphical representation of~\eqref{eq:gssm-system} is given in Figure~\ref{fig:gssm_bare}. 

While several problem settings can be cast within this framework, in this paper, we focus on the setting where the data consist of a collection of related time series \citep{cini2025graph}.
In this setting, natural in many applications, the observed data consist of several time series evolving along a common temporal axis, where each series is associated with a sensor, agent, or virtual entity. Interactions and relations among these entities may arise from, \eg, physical constraints, spatial proximity, or causal relations, which can be represented as graphs and result in correlated time series. 
Accordingly, in this setting, both input and output graphs are defined in terms of these time series~(graph nodes) and the relations~(graph edges) among the entities generating them. We refer to such entities as sensors for simplicity.

\paragraph{Inputs}

\begin{figure}
\centering
\includegraphics[width=.75\textwidth]{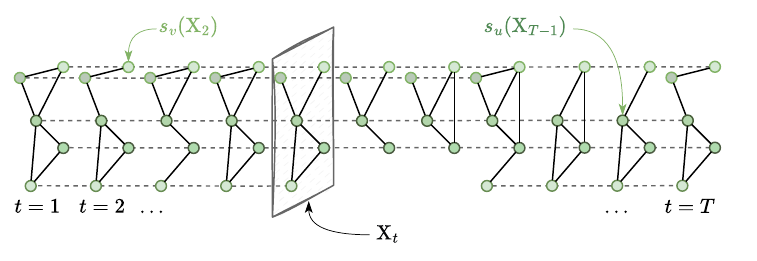}
\caption{An example of a spatio-temporal data over a set $\Vx$ of 5 nodes. Graph $\gx_t$ is given at each step $t$. $\gx_t$ is defined over a node set $V(\gx_t)\subseteq\Vx$ and has node signals $s_v(\gx_t)\in\mathbb R^{d_x}$ associated with each node $v\in V(\gx_t)$.}
\label{fig:dynamic-graph}
\end{figure}

The input graph $\gx_t$ at time $t$ is defined by a finite node set $V(\gx_t)$, associated node attributes $s(\gx_t)=\{s_v(\gx_t)\in\mathbb R^{d_x}\}_{v\in V(\gx_t)}$, and an edge set $E(\gx_t)\subseteq V(\gx_t)\times V(\gx_t)$ encoding known existing relationships. Nodes correspond to, \eg, sensors, while node attributes, or signals, represent the corresponding (possibly multivariate) sensor readings. 
We refer to relations in $E(\gx_t)$ as \emph{spatial}, in reference to the dimensions spanned by sensors.
Node sets, topologies, and node attributes may all change with $t$; $\Vx=\bigcup_t V(\gx_t)$ denotes the finite union of all nodes observed over the time horizon. 
This framework allows for covering many different setups. For example, the above setting includes the standard setup to modeling multivariate time series where: 1) time series are typically assumed to be regularly sampled, 2) sensors are available at all time steps ($V(\gx_t)=\Vx\ \forall t$), and 3) no relational information is available as a prior ($E(\gx_t)=\emptyset\ \forall t$). At the same time, it allows for more advanced setups. 
Examples include cyber-physical systems consisting of interconnected entities ($E(\gx_t)\ne\emptyset\ \forall t$) and social networks in which new nodes and edges appear and disappear over time ($V(\gx_t)$ and $E(\gx_t)$ are dynamic); \eg, see \citep{trivedi2018dyrep, kazemi2020representation}.  
A visual representation of the graph-based representation for the spatio-temporal data sequence $\gx_1,\gx_2,\dots,\gx_t,\dots$ is shown in Figure~\ref{fig:dynamic-graph}.
Finally, exogenous variables and additional attributes can be included as well and encoded in $\gx_t$ for notational simplicity.

\paragraph{Outputs}
Given the input sequence $\{\gx_t\}_t$ with node set $\Vx$, we aim to predict output graph $\gy_t$ at time step $t$. 
Graph $\gy_t$ is defined by its node set $V(\gy_t)$, edge set $E(\gy_t)$ and signals $\vy_t = s(\gy_t)$. 
$\gy_t$ is modeled as the realization of a random attributed graph distributed according to \eqref{eq:gssm-system}.
Moreover, node sets $V(\gy_t)$ and $\Vy=\bigcup_t V(\gy_t)$ do not necessarily correspond to $V(\gx_t)$ or $\Vx$, thereby allowing the modeling of several relevant application settings. %, e.g., when one needs to distinguish the actuators of a system from its perceptive apparatus. 
In many applications, the objective is to forecast the temporal evolution of the system given past observations, with $\vy_t \eqdef \vx_{t+H} = s(\gx_{t+H})$ or $\vy_t\eqdef[\vx_{t+1},\dots,\vx_{t+H}]$, for some integer $H>1$.
In other applications, however, $\gy_t$ might be a scalar or a vector encoding of graph-level quantities.

\paragraph{States}
In state-space models, the distribution $\pr(\gy \,|\, \gx_{\le t},\gh_0)$ of the model output $\gy_t$ reduces to $\pr(\gy\,|\,\gh_{t})$ when current state $\gh_t$ is known; this follows from the conditional independence of $\gy_t$  from the past ($\gx_1,\dots,\gx_{t},\gh_0$), characteristic of state-space models. 
We model the state $\gh_t$ as an unobservable, attributed graph defined by node set $V(\gh_t)$, edge set $E(\gh_t)$, and node state signals $s(\gh_t)$. The state $\gh_t$ is a random variable with distribution $\pr(\gh\,|\,\gh_{t-1},\gx_t)$ conditioned on the previous state $\gh_{t-1}$ and the most recent input $\gx_t$; node set and topology of $\gh_t$ are allowed to vary over time and need not correspond to that of $\gx_t$ or $\gy_t$. This probabilistic, graph-based formulation of the states is a central and novel contribution of the paper.

\medskip
By following the formulation of the data-generating process~\eqref{eq:gssm-system}, we introduce the {Graph State-Space} (GSS) family of predictive models
\begin{equation}\label{eq:gssm-approx}
\begin{cases} 
    \gh_t \sim \pr_\theta(\gh\mid  \gh_{t-1}, \gx_t)
    \\
    \hat \gy_t \sim \pr_\psi(\hat \gy\mid \gh_t),
\end{cases}
\end{equation}
intended to approximate the input--output relation observed in the data generated by~\eqref{eq:gssm-system}.
The model family is parametrized by learnable parameters $\theta$ and $\psi$, governing respectively the stochastic state update $\gh_{t-1}\mapsto \gh_{t}$ and the readout  mapping $\gh_{t}\mapsto \hat \gy_{t}$. 
Note that, with a small abuse of notation, we use the same symbol $\gh_t$ for the states of the system \eqref{eq:gssm-system} and those of the predictive model \eqref{eq:gssm-approx}. However, we emphasize that the inferred states from~\eqref{eq:gssm-approx} do not necessarily need to estimate the unobservable states of the data-generating process~\eqref{eq:gssm-system} directly.

A key contribution of this paper is showing that GSS model parameters can be learned from realizations $\{\gx_t,\gy_t\}_{t=1}^T$ of the data-generating process~\eqref{eq:gssm-system}, by optimizing the predictions $\hat\gy_t$ of $\gy_t$ given the input graphs $\gx_t$, without relying on observed states from~\eqref{eq:gssm-system}. 
In particular, the model is learned to infer graph-structured states, with potentially variable node sets, topology, and attributes, without assuming a specific correspondence between the node sets $\Vh$, $\Vx$, and $\Vy$. The next section elaborates on the relationships between node sets and their implications in more detail, while a discussion about model training follows in Section~\ref{sec:learning-strategy}.

\subsection{Relationships among Input, State, and Output Nodes}
\label{sec:node-relationships}

A central difficulty in generalizing state-space models to graph state representations lies in handling relationships among the node sets of $\gx_t$, $\gh_t$, and $\gy_t$, and correspondences among nodes along the temporal dimension. 
Node sets $V(\gx_t)$ and $V(\gx_{t'})$ might be different but, typically, $V(\gx_t)\cap V(\gx_{t'})\ne \emptyset$ implying some partial correspondence among groups of nodes;
see Figure~\ref{fig:dynamic-graph} for a visual example.
When this is the case, $\vx_{t:T}\eqdef s(\gx_{t:T})$ denotes the tensor $[\vx_t,\dots,\vx_{T-1}]\in\R^{(T-t) \times|\Vx|\times d_x}$ obtained by stacking all graph signals and appropriately masking any missing nodes at each time step. Analogously, sequences of outputs and states are denoted by $\vy_{t:T}\eqdef s(\gy_{t:T})$ and $\vh_{t:T}\eqdef s(\gh_{t:T})$, respectively. 

\begin{figure}
    \begin{subfigure}[b]{\textwidth}
        \includegraphics[width=\textwidth]{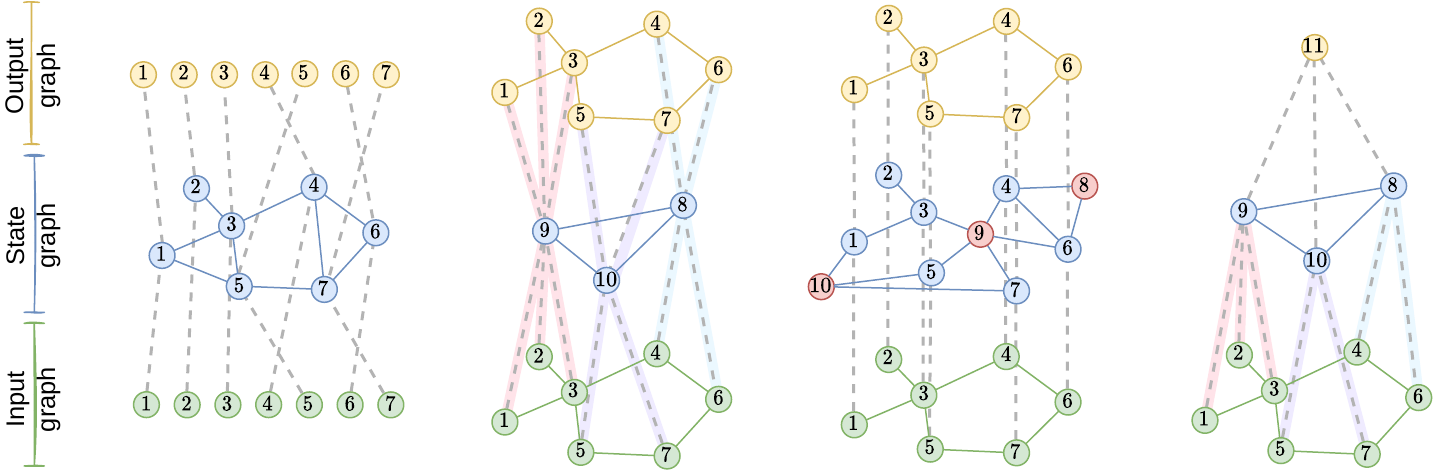}
    \end{subfigure}
    \\
    \hspace*{.05\textwidth}
    \begin{subfigure}[b]{0.25\textwidth}
        \centering
        \rule{\linewidth}{0pt}
        \caption{}
        \label{fig:modalities-nograph}
    \end{subfigure}
    \begin{subfigure}[b]{0.20\textwidth}
        \centering
        \rule{\linewidth}{0pt}
        \caption{}
        \label{fig:modalities-factor-analysis}
    \end{subfigure}
    \begin{subfigure}[b]{0.28\textwidth}
        \centering
        \rule{\linewidth}{0pt}
        \caption{}
        \label{fig:modalities-partial}
    \end{subfigure}
    \begin{subfigure}[b]{0.19\textwidth}
        \centering
        \rule{\linewidth}{0pt}
        \caption{}
        \label{fig:modalities-global}
    \end{subfigure}%
    \caption{Some examples of possible configurations of the input, state, and output graphs and node correspondence for different problem settings described in Section~\ref{sec:node-relationships}. Dashed lines and node numbering encode the correspondence between nodes in sets $\Vx,\Vh$, and $\Vy$.}
    \label{fig:modalities}
\end{figure}

Conversely, although this might be the case in some scenarios, we are not assuming identified nodes among sets $\Vx,\Vh$, and $\Vy$; in particular, we are not assuming a one-to-one correspondence between $\Vx$ and $\Vh$ or $\Vh$ and $\Vy$. 
Indeed, removing this assumption makes the problem both more difficult and interesting. For instance, if $\gy_t$ is a partial observation of $\gh_t$ as in Figure~\ref{fig:modalities-partial}, then $\Vy\subseteq \Vh$. Differently, in a factor analysis setting, the two sets $\Vh$ and $\Vy$ might be disjoint; see Figure~\ref{fig:modalities-factor-analysis}. 
The proposed formulation is very general and encompasses several successful models for time series analysis, some of which are described in Section~\ref{sec:existing-instances}.

Other scenarios are illustrated in Figure~\ref{fig:modalities}. 
For instance, in Figure~\ref{fig:modalities-nograph}, the task involves node-level prediction with relational inference, in which all three sets of nodes are identical, $\Vx = \Vh = \Vy$, and the input graph $\gx_t$ lacks an observable relational structure, which is instead learned directly from the data and incorporated into the state graph $\gh_t$. 
Another scenario, depicted in Figure~\ref{fig:modalities-factor-analysis}, involves learning hidden factors for unsupervised learning, where $\gx_t$ and $\gy_t$ coincide, and one-to-many relationships from $\Vh$ to $\Vx$ allow node signals in $\gh_t$ to represent latent factors driving the observed dynamics. 
Figure~\ref{fig:modalities-partial} can represent a setting of time series forecasting with partial observation, where the node correspondence among the three graphs is incomplete, and some nodes remain hidden, leading to $\Vx = \Vy \subset \Vh$; in this context, $\gy_t=\gx_{t+1}$, and while $\gh_t$ possesses its own graph structure, additional relations are captured through $\gh_t$. 
Finally, Figure~\ref{fig:modalities-global} presents a graph-level prediction scenario, in which information from all sensors in $\Vx$ is transformed into a more compact set of nodes in $\Vh$, producing a single, graph-level output.

Note that within the proposed GSS framework, the relational structure associated with the edges $E(\gh_t)$, and the mappings $V(\gx_t)\to V(\gh_t)$ and $V(\gh_t)\to V(\gy_t)$ might be entirely or partially latent, and this sets our approach apart from the existing literature.

\subsection{Implementing a Graph State-Space Model}
\label{sec:implementation}

We now describe a general encoder--decoder architecture implementing the stochastic graph state-space model in~\eqref{eq:gssm-approx}. The goal is to provide a unifying and modular architecture that accommodates a wide range of application setups and tasks.

The architecture is organized into three conceptual blocks:
(i) an \emph{input encoder} mapping the observed graph $\gx_t$ to the latent state space $\mathcal H$,
(ii) a \emph{state transition} to update the state graph over time, and
(iii) a \emph{readout} mapping latent states to outputs.
While we present an architecture based on graph pooling and message passing, the proposed formulation serves as a blueprint rather than prescribing a unique implementation. The architecture can be further tailored depending on the application.
Section~\ref{sec:experiments} later presents several model configurations implementing different node relationships.

\subsubsection{Input Encoder}
\label{sec:input-encoder}

The input encoder maps the input graph $\gx_t$, defined over the input node set $\Vx$, to a representation aligned with the latent state nodes $\Vh$.
This mapping accounts for a potential mismatch between the two node sets and supports dimensionality reduction or augmentation.

We adopt the Select--Reduce--Connect (SRC) framework for graph pooling \citep{grattarola2024understanding}, which provides a general abstraction for learnable graph-to-graph mappings.
Within this framework, the \emph{select} operator defines a (soft or hard) correspondence between nodes of an input graph and a target graph (here in $\mathcal X$ and $\mathcal H$) through an affiliation matrix $\vect S$.
Then, node features and connectivity of the target graph are computed from the input graph and matrix $\vect S$ via \emph{reduce} and \emph{connect} operators, respectively.

In practice, following the SRC framework, the input encoder creates a correspondence between nodes in $V(\gx_t)$ and a subset $V_{\xth}\subseteq\Vh$ of state nodes determined by affiliation matrix
\begin{equation}
\vect S_t = \textsc{Select}(\gx_t;\theta) \in \R^{|V_{\xth}|\times |V(\gx_t)|}.
\end{equation}
Different parameterizations of the select operator have been used in the literature. For instance, it can be implemented as an MLP \citep[\eg, see][]{bianchi2020spectral}, exploit the graph topology \citep[as in][]{ying2018hierarchical}, or be fixed by design when specific node correspondences are required.
Though defined as input-dependent to account for a time-varying matrix $\vect S_t$, the select operator can also be defined to learn static relationships between $\Vx$ and $\Vh$.

Given $\vect S_t$, the \emph{reduce} operator aggregates node signals from $\gx_t$ and maps them into latent representations as
\begin{equation}
\vxth_t
= \textsc{Reduce}(\gx_t,\vect S_t;\theta) \in \R^{|V_{\xth}|\times d_{\xth}}
\end{equation}
for some feature dimension $d_{\xth}$.
A common choice is to use the linear aggregation  $\vxth_t=\vect S_t \vx_t$.
The resulting node features $\vxth_t$ are used to condition the state transition described in the next section, where the remaining step of SRC, the connect operation, is carried out.

\subsubsection{State Transition}
\label{sec:state-transition}

The state transition module updates both the topology and node representations of the state graph $\gh_{t-1}$ to produce $\gh_t$. Unlike input graphs, state graphs are random entities; therefore, the layers presented here model graph distributions.

The connectivity of the next state $\gh_t$ is generated by the \emph{connect} operator, which defines a learnable probability distribution for the edge set $E_t=E(\gh_t)\subseteq \Vh\times\Vh$,
\begin{equation}
E_t \sim \pr_\theta\left(E \mid \vxth_t, \gh_{t-1}\right).
\end{equation}
In this work, we instantiate it as a collection of Bernoulli distributions parameterized by a matrix of logits $\Phi_t=\Phi(\vxth_t,\gh_{t-1};\theta)\in\R^{|\Vh|\times|\Vh|}$, so that each edge $(i,j)$ is distributed as $\text{Bernoulli}(\sigma([\Phi_t]_{i,j}))$. 
This formulation allows the latent topology to evolve over time, given the current inputs and the previous state. 
Examples include an input- and state-independent matrix $\Phi$ of free parameters in $\theta$, or definitions where $[\Phi_t]_{i,j}=\vh_{t-1}(i)^\top\vh_{t-1}(j)$ for $(i,j)\in E(\gh_{t-1})$ and zero otherwise.

For notational convenience, we define $\vhth_{t-1}\eqdef\left[\vh_{t-1} \,\middle\lVert\, \vxth_t\right],$ the concatenation along the feature dimension of $\vh_{t-1}=s(\gh_{t-1})$ and $\vxth_t$.
Row alignment between $\vxth_t$ and $\vh_{t-1}$ follows their indexing in $\Vh$, with zero-padding when a node is missing from either representation.

Given the sampled edge set $E_t$, node representations are updated via message passing,
\begin{equation}
\vh_t = \textsc{MP}(\vhth_{t-1},E_t;\theta) \in \R^{|V_t|\times d_h},
\end{equation}
where $V_t$ is defined as the union of $V(\gh_{t-1})$, $V_{\xth}$, and the nodes incident to edges in $E_t$.
The message-passing operator can combine standard GNN operations with probabilistic layers, defining node distributions that depend more directly on the input, the topology $E_t$, and the previous state.
For instance, the conditional distribution of $\vh_t$ given $\vhth_{t-1}$ and $E_t$ can be modeled as the push-forward of a base noise distribution $\pr(\vect \nu)$ through a GNN,
\begin{equation}\label{eq:gnn-reparam}
\vh_t = \textsc{GNN}([\vhth_{t-1}\,\lVert\,\vect \nu],E_t;\theta),
\qquad \vect \nu \sim \pr(\vect \nu).
\end{equation}
An alternative is that of $\vh_t$ following a parametric distribution $\pr(\vh;\phi)$ whose parameters $\phi$ are produced by a GNN,
$\phi=\textsc{GNN}(\vhth_{t-1},E_t;\theta)$.

The resulting state graph at time $t$ is $\gh_t \eqdef (V_t,E_t,\vh_t)$, whose distribution corresponds to $\pr_\theta(\gh\mid\gh_{t-1},\gx_t)$ in~\eqref{eq:gssm-approx}.

\subsubsection{Readout}
\label{sec:readout}

The readout maps the state graph $\gh_t$ to the output as
\begin{equation}
\hat\gy_t = \textsc{Readout}(\gh_t;\psi) 
\end{equation}
and its architecture depends on whether the task is graph-level or node-level.
For graph-level outputs, i.e., where $|\Vy|=1$, a global pooling operator is applied to $\gh_t$.
For tasks associated with the nodes of $\gx_t$, such as spatio-temporal forecasting, the readout must lift representations from $\Vh$ back to $\Vx$.
This lifting, or unpooling, operation can be implemented using the pseudo-inverse of the affiliation matrix, $\vect S_t^+ s(\gh_t)$, which redistributes latent information to the original node set \citep{grattarola2024understanding}.
In both cases, the predictive distribution $\pr_\psi(\hat\gy\,\mid\,\gh_t)$ can be modeled via probabilistic readouts implementing strategies similar to those described at the end of Section~\ref{sec:state-transition}.
In the following, as a reference approach, we consider a readout based on the reparametrization trick, such that
\begin{equation}
\hat\gy_t = g_\psi(\gh_t, \vect \varepsilon) \qquad\text{ with } \vect \varepsilon \sim \pr(\vect \varepsilon),
\end{equation}
for some function $g_\psi$ differentiable in $\psi$.
More advanced readout mechanisms can be devised for graph generative tasks~\citep{guo2022systematic}.

\section{Learning Strategy}\label{sec:learning-strategy}

A GSS model involves latent and discrete variables, which pose non-trivial challenges for gradient-based optimization. 
This section shows that parameter vectors $\psi$ and $\theta$ of the model can nevertheless be effectively learned from a training set $\{\gx_t,\gy_t\}_{t=1}^T$. 
In particular, we show how to train GSS models end-to-end by minimizing an objective function $\mathcal L_t(\psi,\theta)$ defined on the model prediction $\hat \gy_t$, without access to the latent state of the system. 

For point predictions, the objective function can be formulated as
\begin{align}\label{eq:pp-loss}
\mathcal L_t(\psi,\theta) &= \E_{\gy_t}\left[\ell(\gy_t,T[\hat \gy_t])\right],
\end{align}
where $\ell$ is a loss function that assesses the discrepancy between the target $\gy_t$ and a point prediction $T[\hat \gy_t]$.
% \New{
As $\hat \gy_t$ is a random graph, point predictions $T[\hat \gy_t]$ are obtained from the model's predictive distribution
\begin{equation}\label{eq:P-psi-theta}
    \pr_{\theta,\psi}^t = \pr_{\theta,\psi}(\hat \gy_t \mid \gx_t,\dots,\gx_1,\gh_0)
\end{equation}
of $\hat \gy_t$ given past observations, or $\pr_{\theta,\psi}(\hat \gy_t \mid \gh_t)$, by exploiting 
the conditional independence from the past given the current state.
Typical examples of $T$ include 1) the expected value, yielding---for vector outputs---the optimal $L^2$-norm prediction, 2) the mode of $\pr_{\theta,\psi}^t$, which provides the maximum likelihood estimate, and 3) the quantile at a given level $\alpha$, which can be used to provide confidence regions. 
For graph data, generalizations exist. For instance, the expected value can be expressed as a barycenter of the distribution, defined as the minimizer of the Fr\'echet function $\mathcal F(\gy)=\mathbb E_{\hat \gy\sim \pr_{\theta,\psi}^t}[d(\gy,\hat\gy)]$ with respect to a given graph distance $d$ \citep{frechet1948elements,jain2016statistical}.
Examples of loss $\ell$ include the $L^2$- and $L^1$-norms, corresponding to node-level mean squared error (MSE) and mean absolute error (MAE), respectively, as well as the so-called pinball loss used in quantile regression~\citep{koenker2001quantile}.
Another example, which we will consider as a reference case, is given by:
\begin{align}\label{eq:expected-loss}
\mathcal L_t(\psi,\theta) &=\E_{\gy_t}\left[\E_{\hat\gy_t}\left[\ell(\gy_t,\hat \gy_t)\right]\right].
\end{align}
where the objective function averages the loss $\ell$ across realizations of $\hat \gy_t$. 

More generally, probabilistic models can be learned via objective functions defined directly on the predictive distribution $\pr^t_{\theta,\psi}(\hat \gy_t)$. Examples include the log-likelihood of the target $\gy_t$ under the distribution of $\hat \gy_t$, or the continuous ranked probability score \citep[CRPS,][]{gneiting2007strictly}, both of which generalize to graph data \citep{rizzo2016energy}. 
Although certain objective functions are known to have stronger theoretical guarantees for learning probabilistic models with latent variables than others \citep{manenti2025learning}, in this section, we focus on~\eqref{eq:expected-loss} to present the learning procedure more clearly.

Optimizing~\eqref{eq:expected-loss} with gradient descent requires estimating gradients $\nabla_\psi\E_{\hat\gy_t\sim \pr_{\theta,\psi}^t}\left[\ell(\gy_t,\hat \gy_t)\right]$  and
$\nabla_\theta\E_{\hat\gy_t\sim \pr_{\theta,\psi}^t}\left[\ell(\gy_t,\hat \gy_t)\right]$. 
Computing analytic solutions for these gradients is generally intractable and impractical, even with automatic differentiation tools.
We address this challenge by adopting Monte Carlo estimators to approximate the above gradients. In particular, we use estimators based on the reparametrization trick and the score-based reformulation. Alternative optimization strategies may be applicable depending on the specific model implementation and loss function $\ell$.

\subsection{Readout Gradient Estimation}
Given state graph $\gh_t$, the randomness of the readout output $\hat \gy_t=g_\psi(\gh_t,\vect \varepsilon)$ is entirely due to the auxiliary noise variable $\vect \varepsilon$. 
Reparametrizing samples from the predictive distribution $\pr_{\psi}(\hat \gy\mid \gh_t)$ 
yields
\begin{align}\label{eq:dec:gradient-estimation-RT}
    \nabla_{\psi}\E_{\hat\gy_t|\gh_t}\left[\ell(\gy_t,\hat \gy_t)\right]
  =\nabla_{\psi}\E_{\vect\varepsilon}\left[\ell(\gy_t,g_\psi(\gh_t, \vect\varepsilon))\right]
  =\E_{\vect\varepsilon}\left[\nabla_{\psi}\ell(\gy_t,g_\psi(\gh_t, \vect\varepsilon))\right]
\end{align}
which admits the Monte Carlo estimator
\begin{equation}\label{eq:dec:gradient-estimation-MC}
\frac{1}{M} \sum_{m=1}^M \nabla_{\psi}\ell\left(\gy_t, g_\psi\left(\gh_t, \vect\varepsilon^m\right)\right).
\end{equation}
given $M$ \iid samples $\vect\varepsilon^m\sim \pr(\vect\varepsilon)$. The reparametrization trick thus provides a simple and general gradient estimator that does not require additional assumptions on the loss function or on the form of the predictive distribution.

\subsection{State Transition Gradient Estimation}

To tackle the estimation of gradient $\nabla_\theta\E_{\hat\gy_t\sim \pr_{\theta,\psi}^t}\left[\ell(\gy_t,\hat \gy_t)\right]$, we isolate the dependence on the parameter vector $\theta$
\begin{align}
\nabla_\theta \E_{\hat\gy_t}\left[\ell(\gy_t,\hat\gy_t)\right]
  = \nabla_\theta \E_{\gh_t}\left[\E_{\hat\gy_t|\gh_t}\left[\ell(\gy_t,\hat\gy_t)\right]\right]
  = \E_{\vect \varepsilon}\left[ \nabla_\theta \E_{\gh_t}\left[\ell(\gy_t,\hat\gy_t)\right]\right].
\end{align}
We therefore focus on the term
\begin{equation}
    \nabla_\theta \E_{\gh_t}\left[\ell(\gy_t,\hat\gy_t)\right] = 
    \nabla_\theta \E_{E_t}\left[\E_{\vh_t|E_t}\left[\ell(\gy_t,\hat\gy_t)\right]\right],
\end{equation}
where the distribution of both $E_t$ and $\vh_t|E_t$ depend on $\theta$.

Let $\theta_{s}$ and $\theta_e$ denote the subsets of parameters in $\theta$ governing the distribution of node signals $\vh_t\mid E_t$ and topology $E_t$, respectively.
If $\vh_t$ is modeled using the reparametrization trick, \eg, with predefined noise $\vect\nu$ as in~\eqref{eq:gnn-reparam}, the gradient with respect to ${\theta_s}$ can be written as
\begin{align}
\nabla_{\theta_s} \E_{\gh_t}\left[\ell(\gy_t,\hat\gy_t)\right] =
\E_{E_t}\left[\E_{\vect\nu}\left[\nabla_{\theta_s} \ell(\gy_t,\hat\gy_t)\right]\right],
\end{align}
and can be estimated analogously to the readout parameters $\psi$. 

The parameters $\theta_e$, instead, require additional care, since $E_t$ is a \emph{discrete} random variable. Instead of the reparametrization trick, we resort to a score-based gradient estimator \citep{mohamed2020monte}. Specifically, we write
\begin{align}\label{eq:enc:gradient-estimation-score}
\nabla_{\theta_e}\E_{E_t}\left[\ell(\gy_t,\hat\gy_t)\right] 
= 
\E_{E_t}\left[\ell(\gy_t,\hat\gy_t)\;\nabla_{\theta_e}\log\,\pr_{\theta_e}^t(E_t)\right] 
\end{align}
where $\pr_{\theta_e}^t$ denotes the likelihood of $E_t$ and $\nabla_{\theta_e}\log\,\pr_{\theta_e}^t(E_t)$ is usually called score function. In this form, the gradient in~\eqref{eq:enc:gradient-estimation-score} requires differentiating only the score function with respect to ${\theta_e}$, enabling the Monte Carlo estimator
\begin{align}
\frac{1}{M} \sum_{m=1}^M \ell\left(\gy_t, \hat\gy_t|E_t^m \right) \nabla_{\theta_e}\log \pr_{\theta_e}^t(E_t^m)
\end{align}
where $\{E_t^m\}_{m=1}^M$ are \iid samples drawn from $\pr_{\theta_e}^t$. % defined in \eqref{eq:enc:con}.

The use of score-based estimators allows the edges of the random state graph $\gh_t$ to be treated as discrete objects during both training and inference, thereby keeping all message-passing operations sparse and computationally efficient~\citep{cini2023sparse}.

\section{Casting Existing Work as Graph State-Space Models}
\label{sec:existing-instances}

In this section, we review representative methods from the literature related to relational state-space representations and reinterpret them under the formalism introduced in Section~\ref{sec:model-formulation}.
This perspective highlights common structural components across otherwise disparate approaches and clarifies their relationship to the proposed framework.

We note that, to the best of our knowledge, the present work is the first to introduce a state-space model with stochastic graph-valued latent states whose distribution is learned end-to-end from data in conjunction with a downstream prediction task, and whose topology is allowed to differ from that of the input graphs ($E(\gh_t)\neq E(\gx_t)$).

\subsection{Deep Factor Models with Random Effects}

\citet{wang2019deep} propose a deep factor model for time series forecasting ($\Vx=\Vy$).
The model introduces a set of $K<|\Vx|$ global latent factors that augment the observed nodes, yielding a state node set $\Vh\supset\Vx$.

The state graph topology connects each observed node to the global factors. State transitions are performed independently at the node level and can be summarized as
\begin{equation}
    \vh_{t,u} \sim \pr_\theta(\vh_u \mid \vh_{t-1,u}, \vx_t),
    \qquad \forall\,u\in\Vh.
\end{equation}
The readout relates the latent state to the outputs via
\begin{equation}
    \hat\vy_{t,v} \sim \pr(\hat\vy_v \mid \gh_t,\psi),
    \qquad \forall\,v\in\Vy.
\end{equation}

\subsection{Clustering-based Aggregate Forecasting}

The nonlinear method proposed by \citet{cini2020cluster} addresses the problem of predicting an aggregate quantity
$\vy_t=\sum_{v\in\Vx}\vx_{t,v}\in\R$
from a collection $\Vx$ of smart meters. No relational information among sensors is assumed a priori.

The model learns a set $\Vh$ of clusters and a binary affiliation matrix
$\vect S\in\{0,1\}^{|\Vh|\times|\Vx|}$ used to aggregate input signals.
The aggregated signals are then used to update the latent state via
\begin{equation}
    \vh_t = \text{RNN}_\theta(\vh_{t-1}, \vect S\,\vx_t).
\end{equation}
The readout produces the final prediction by applying the same function $f_\psi$ to each latent node and summing the results,
\begin{equation}
    \hat\vy_t = \sum_{u\in\Vh} f_\psi(\vh_{t,u}).
\end{equation}

Under the proposed framework, $\vect S$ corresponds to the output of a select operator, while the readout implements a specific instance of global pooling; see also Figure~\ref{fig:modalities-global}.
The model is deterministic and does not account for uncertainty in either the clustering assignments or the latent state dynamics.

\subsection{Topology Identification from Partial Observations}

\citet{coutino2020state} consider a system composed of a set $\Vh$ of interacting agents with states $\vh_t$ and address the problem of identifying the relational structure of the system, an edge set $E_{\gh}\subseteq \Vh\times\Vh$, from partial observations.
Specifically, only a subset $\Vy\subset\Vh$ of agents is observed, yielding a setting closely related to that illustrated in Figure~\ref{fig:modalities-partial}.

The graph topology $E_{\gh}$ is assumed to be static, \ie, $E(\gh_t)=E_{\gh}$ for all $t$, and is inferred by postulating a linear state transition model of the form
\begin{equation}
    \vh_t = A_\theta(E_{\gh})\,\vh_{t-1} + B_\theta\,\vx_t + \vect e_t,
    \qquad \vect e_t \sim \pr_{\vect e},
\end{equation}
where $A_\theta(E_{\gh})$ is a trainable matrix-valued function of the graph topology and
$B_\theta\in\R^{|\Vh|\times|\Vx|}$ acts as an affiliation matrix relating input signals $\vx_t$ to $\vh_{t}$.
In terms of our framework, $B_\theta$ can be interpreted as a fixed instance of the select operator.

The readout function is also linear and given by
\begin{equation}
    \hat \vy_t = C_\psi\,\vh_t + D_\psi\,\vx_t + \vect e_t',
    \qquad \vect e_t' \sim \pr_{\vect e'}.
\end{equation}

\subsection{Adaptive Graph Convolutional Recurrent Network}\label{sec:cast-agcrn}

The Adaptive Graph Convolutional Recurrent Network (AGCRN) introduced by \citet{bai2020adaptive} addresses spatio-temporal forecasting tasks using only time series data as input; no graph structure is assumed for the input signal, \ie $\Vy=\Vx$ and $E(\gx_t)=\emptyset$.

AGCRN learns a static latent graph topology $E_{\gh}$ that captures pairwise relations among nodes and is shared across time.
The weighted adjacency matrix associated with $E_{\gh}$ is factorized as $\text{softmax}(\text{ReLU}(\vect E \, \vect E^\top))$, with $\vect E \in \R^{|\Vx|\times d_E}$ a matrix of learnable free parameters.
This learned topology is then used by a graph neural network to update the hidden state.
Within our framework, AGCRN can be written as
\begin{equation}
\begin{cases}
    \vh_t = \textsc{MP}_\theta\left([\vx_t \lVert \vh_{t-1}],E_{\gh}\right),\\
    \hat\vy_t = f_\psi(\vh_t).
\end{cases}
\end{equation}

AGCRN thus corresponds to a deterministic graph state-space model with a learned but time-invariant state topology and a message-passing-based state transition.

\subsection{Probabilistic Spatio-Temporal Forecasting}
\citet{pal2021rnn} study a probabilistic spatio-temporal forecasting problem in which the input signal $\vx_t$ is associated with a known and time-invariant graph topology $E_{\gx}$.
The node sets coincide across input, state, and output graphs, \ie $\Vx=\Vh=\Vy$.

Cast within our framework, their model can be expressed as the state-space system
\begin{equation}
\begin{cases}
    \vh_t = f_{\theta}\left(\vh_{t-1},\vx_t,E_{\gx},\vect e_t\right),\\
    \vy_t = f_{\psi}\left(\vh_t,E_{\gx},\vect e'_t\right),
\end{cases}
\end{equation}
where $\vect e_t\sim \pr_{\theta}(\vect e\mid\vh_{t-1})$ and
$\vect e'_t\sim \pr_{\psi}(\vect e'\mid\vh_t)$ model stochasticity in the state transition and readout, respectively.
Besides the input graph topology $E_{\gx}$, the model can exploit learned relations $E_{\gh}$ for the state as done by \citet{bai2020adaptive}, and described in previous section.

\section{Experimental Setup and Validation}
\label{sec:experiments}

This section empirically validates both the soundness of the proposed Graph State-Space (GSS) modeling framework and the effectiveness and interpretability of its model instantiations.
Since the primary goal of the paper is to methodologically advance state-space modeling and deep learning toward more problem-agnostic architectures with interpretable model components, the experimental analysis is designed to assess the following claims:
\begin{enumerate}[label={(\roman*)}]
\item The parameters of GSS models can be learned effectively end-to-end to solve the target prediction tasks, despite the challenges posed by learning discrete structures and by the multiple components of the architecture introduced in Section~\ref{sec:implementation}.
\item The same general framework supports the implementation of different architectures tailored to the requirements of a given task, beyond the examples illustrated in Figure~\ref{fig:modalities} and Section~\ref{sec:existing-instances}.
In particular, GSS models are shown to recover ground-truth topological information, form meaningful node clusters, and promote long-range information propagation over graphs when required by the task.
\item The generality and advantages of GSS models do not come at the expense of predictive performance, with accuracy comparable to or exceeding that of existing GSS-based approaches and competitive with other classes of methods.
\end{enumerate}
Experiments are conducted on two datasets: a synthetic benchmark that provides access to ground-truth relational structure for validation purposes, and a real-world dataset drawn from energy production management, used to demonstrate the applicability of the proposed framework in a practical setting.

\begin{table}
\centering
\resizebox{\textwidth}{!}{%
\begin{tabular}{rrccccc}
\toprule
&                & \textbf{Input} & \textbf{Input-state} & \textbf{State} & \textbf{State} & \textbf{State-output} \\
&\textbf{Model}  & \textbf{topology} & \textbf{node corresp.} & \textbf{nodes} & \textbf{topology} & \textbf{node corresp.} \\
&                & $E(\gx_t)$ &  $V(\gx_t)\leftrightarrow V(\gh_t)$ & $V(\gh_t)$ & $E(\gh_t)$ & $V(\gh_t)\leftrightarrow V(\gy_t)$ \\
\midrule
\multirow{6}{0em}{\rotatebox[origin=c]{90}{GSS Baselines}}
&{RNN}      & $\emptyset$       & identity                & $V(\gx_t)$     & $\emptyset$      & identity  \\
&{fc-RNN}   & $\emptyset$       & fully connected         & $|V(\gh_t)| = 1$ & $\emptyset$    & fully connected  \\
&{stt-STGNN}& encoder           & identity                & $V(\gx_t)$     & $\emptyset$      & identity  \\
&t\&s-STGNN & state tr.         & identity                & $V(\gx_t)$     & $E(\gx_t)$       & identity \\
&{tts-STGNN}& readout           & identity                & $V(\gx_t)$     & $E(\gx_t)$       & identity  \\
&DCRNN      & state tr.         & identity                & $V(\gx_t)$     & $E(\gx_t)$       & identity  \\
\midrule
\multirow{5}{0em}{\rotatebox[origin=c]{90}{GSS Models}}
&{Id-GSS}   & $\emptyset$       & identity                & $V(\gx_t)$              & learned          & identity  \\
&{Ext-GSS}  & $\emptyset$       & partial identity        & $V(\gh_t)\supset V(\gx_t)$ & learned       & partial identity  \\
&{Pool-GSS} & $\emptyset$       & learned pooling         & $|V(\gh_t)|<|V(\gx_t)|$ & learned          & lifting  \\
&{Hub-GSS}  & encoder           & learned pooling         & $|V(\gh_t)|<|V(\gx_t)|$ & learned          & lifting  \\
&AGCRN      & $\emptyset$       & identity                & $V(\gx_t)$              & learned          & identity  \\
\bottomrule
\end{tabular}%
}
\caption{Main characteristics of the considered models. Symbol $\emptyset$ means no topological information, while ``encoder'', ``state tr.'', and ``readout'' indicate where the topology is processed. ``fully connected'' indicates that all nodes are related to all nodes through a dense neural layer, and ``lifting'' that the affiliation matrix from the (learned) pooling operator is used.}
\label{tab:models}
\end{table}

\subsection{Models}

We evaluate the proposed GSS framework by comparing its forecasting performance against a diverse set of recurrent and spatio-temporal neural architectures.
The comparison includes several instantiations of the GSS framework, designed to explore different modeling choices, as well as representative baseline methods from the literature.

We first consider a set of GSS models that differ in how latent states are structured and how relational information is handled.
The \textbf{Id-GSS} model represents the most straightforward realization of the framework: input and state nodes are in one-to-one correspondence (identity), no relational information is provided at the input, and the state graph topology is learned entirely from data. This setting closely resembles the one illustrated in Figure~\ref{fig:modalities-nograph}.
In \textbf{Pool-GSS}, the state graph contains fewer nodes than the input, following a setting similar to factor-analysis as in Figure~\ref{fig:modalities-factor-analysis}.
Here, no topological information is assumed as input. Instead, both the mapping from input nodes to state nodes and the latent state topology are learned jointly. From $V(\gh_t)$ to $V(\gy_t)$, the affiliation matrix $\vect S$ learned from the pooling from $\gx_t$ to $\gh_t$ is also used to lift $\gh_t$ to $\gy_t$.
The \textbf{Ext-GSS} model extends id-GSS by augmenting the state with additional latent nodes.
This configuration mirrors the partially observed scenario in Figure~\ref{fig:modalities-partial}. 
The \textbf{Hub-GSS} model builds on Pool-GSS by additionally allowing relational information at the input level, which is processed by the input encoder while still allowing the state graph to be learned (and possibly evolve) freely.
These models are reported in Table~\ref{tab:models} under the ``GSS models'' block.

To assess the benefits of explicitly learning relational structure, we compare these GSS configurations with several baseline architectures (``GSS baselines'' in Table~\ref{tab:models}).
These include a standard \textbf{RNN} operating on independent input nodes, as well as an \textbf{fc-RNN} in which a fully connected encoding layer maps the input to a vector state space, enabling interactions across nodes.
Following the nomenclature in \citep{cini2025graph}, we also consider different spatio-temporal graph neural networks that exploit a given, fixed topology: the \textbf{stt-STGNN}, which applies spatial processing in the input encoder only, before the temporal processing; the \textbf{t\&s-STGNN}, where message passing is performed at every state update with respect to input topology; and the \textbf{tts-STGNN}, which incorporates spatial dependencies only at the readout, to provide predictions. 
While all tge above models are instances of the GSS framework, the models explicitly labeled as GSS adhere more closely to its underlying philosophy by learning relational structure as an integral part of the latent state.

Finally, to provide a broader reference across model families, we include several widely used architectures from the literature.
These comprise \textbf{AGCRN} \citep{bai2020adaptive} discussed also in Section~\ref{sec:cast-agcrn}, which learns a static latent topology from time series data; \textbf{DCRNN} \citep{li2018diffusion}, a recurrent spatio-temporal GNN; \textbf{GraphWaveNet} \citep{wu2019graph}, a convolutional architecture that combines a given graph with a learned one; and \textbf{tts-Transf}, a time-then-space transformer that integrates temporal self-attention with graph-based spatial modeling.
Implementation details and hyperparameters are reported in the Appendix~\ref{app:hparams}.

\subsection{Effective Model Training}

\subsubsection{Synthetic dataset}

\begin{figure}
    \centering
    \includegraphics[width=.7\columnwidth,clip,trim={2cm 2cm 2cm 2cm}]{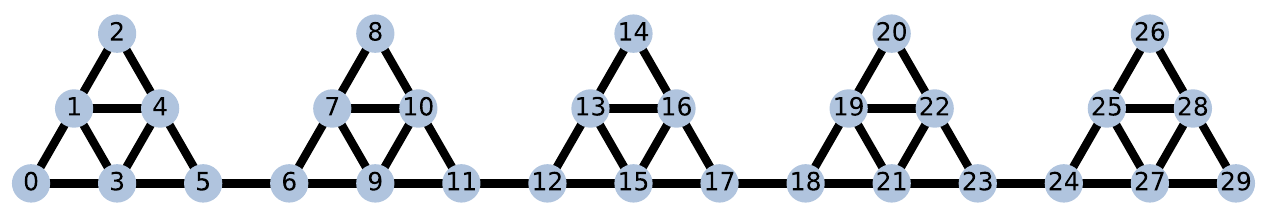}
    \caption{Ground-truth graph topology underlying the GPVAR datasets. Node indices shown here are consistently used in all subsequent figures.}
    \label{fig:gpvar}
\end{figure}

As a synthetic benchmark, we consider the GPVAR dataset \citep{zambon2022aztest}, which features coupled node-level dynamics that can be accurately predicted only by exploiting appropriate relational modeling.
The GPVAR signal $\vect z_{t,v}$ at time $t$ and node $v$ is generated by a polynomial spatio-temporal filter of the form
\begin{equation}\label{eq:gpvar}
\vect z_t = \textrm{tanh}\left(\sum_{l=0}^L\sum_{q=1}^Q \Theta_{l,q}\,\vect A^l \vect z_{t-q}\right) + \varepsilon_t
\qquad \in \R^N.
\end{equation}
Here, $L$ and $Q$ denote the spatial and temporal orders of the filter, respectively, and
$\Theta\in\R^{(L+1)\times Q}$ is a matrix of filter coefficients.
The process evolves according to a fixed graph topology encoded by the adjacency matrix
$\vect A\in\{0,1\}^{N\times N}$.
The noise term $\varepsilon_t$ is drawn independently from a Gaussian distribution
$\mathcal N(\vect 0,\sigma^2\vect I)$.
The process is initialized as $\vect z_{1-q}=\varepsilon_{1-q}$ for $q=1,\dots,Q$.
In the experiments, $\Theta$ is set to $[[5,2],[-4,6],[-1,0]]^\top$, the noise standard deviation to $\sigma=0.4$, and the sequence length to $T=30k$.
The underlying graph contains $N=30$ nodes organized into $5$ communities, as illustrated in Figure~\ref{fig:gpvar}. 

We consider a one-step-ahead forecasting task, where the input signal is
$\vx_t=\vect z_t$, relational information $E(\gx_t)=E_{\gx}$ is provided as input only to some models, and the target $\gy_t$ is $s(\gy_t)=\vx_{t+1}$ with $E(\gy_t)=\emptyset$.
Predictions are produced from windows of $W=9$ past inputs and accuracy is assessed using the mean absolute error (MAE).
The data are divided into 70\%/10\%/20\% splits for training, validation, and testing, respectively.

\newcommand{\sbscr}{\textsubscript}
\newcommand{\rgo}{\cellcolor{green!15}}
\newcommand{\rme}{\cellcolor{yellow!30}}
\newcommand{\rba}{\cellcolor{red!15}}
\begin{table}
\centering
\begin{tabular}{rrccccc}
\toprule
&\multirow{2}{*}{\textbf{Model}} & \multicolumn{2}{c}{\textbf{Relations}} & \multirow{2}{*}{\textbf{Test MAE}}  & \multicolumn{2}{c}{\textbf{AZ-test}}\\
& & {G.T.} & {Learned} & & {Statistic} & {p-value} \\
\midrule
&Analytical opt. & --- & --- & 0.319 & --- & --- \\
\midrule
\multirow{6}{0em}{\rotatebox[origin=c]{90}{GSS Baselines}}
&RNN & no & no & 0.550\textsubscript{$\pm$0.000} & 29.504\textsubscript{$\pm$1.492} & 0.000\textsubscript{$\pm$0.000} \\
&fc-RNN & no & no & 0.431\textsubscript{$\pm$0.004} & 11.642\textsubscript{$\pm$1.534} & 0.000\textsubscript{$\pm$0.000} \\
&tts-STGNN & yes & no & \rgo0.320\textsubscript{$\pm$0.000} & -0.414\textsubscript{$\pm$0.280} & 0.690\textsubscript{$\pm$0.195} \\
&stt-STGNN & yes & no & \rgo0.322\textsubscript{$\pm$0.001} & -0.711\textsubscript{$\pm$0.762} & 0.416\textsubscript{$\pm$0.253} \\
&t\&s-STGNN & yes & no & \rgo0.322\textsubscript{$\pm$0.000} & -0.178\textsubscript{$\pm$0.609} & 0.632\textsubscript{$\pm$0.226} \\
&DCRNN & yes & no & \rgo0.327\textsubscript{$\pm$0.001} & -0.318\textsubscript{$\pm$0.319} & 0.726\textsubscript{$\pm$0.178} \\
\midrule
\multirow{5}{0em}{\rotatebox[origin=c]{90}{\small GSS Models}}
&Id-GSS & no & yes & \rgo0.331\textsubscript{$\pm$0.004} & -0.008\textsubscript{$\pm$0.745} & 0.609\textsubscript{$\pm$0.276} \\
&Ext-GSS & no & yes & \rgo0.332\textsubscript{$\pm$0.007} & -0.192\textsubscript{$\pm$0.447} & 0.757\textsubscript{$\pm$0.237} \\
&Pool-GSS & no & yes & 0.435\textsubscript{$\pm$0.001} & -1.269\textsubscript{$\pm$3.865} & 0.123\textsubscript{$\pm$0.283} \\
&Hub-GSS & yes & yes & 0.397\textsubscript{$\pm$0.002} & -1.140\textsubscript{$\pm$2.155} & 0.360\textsubscript{$\pm$0.415} \\
&AGCRN & no & yes & 0.370\textsubscript{$\pm$0.009} & 4.136\textsubscript{$\pm$1.530} & 0.006\textsubscript{$\pm$0.011} \\
\bottomrule
\end{tabular}
\caption{Results on 1-step ahead forecasting. Reported values are averages of 10 independent runs with standard deviation reported subscripted; values reported as $0.000$ are intended as $<0.001$. Statistics and p-values and AZ-test are also reported. Prediction errors that are within 5\% from the analytical optimal value are highlighted in green. ``G.T.'' column indicates whether the models use the ground-truth graph of Figure~\ref{fig:gpvar} and ``Learned'' if they learn relational information from data.} 
\label{tab:gpvar}
\end{table}

\begin{figure}
    \includegraphics[width=\textwidth]{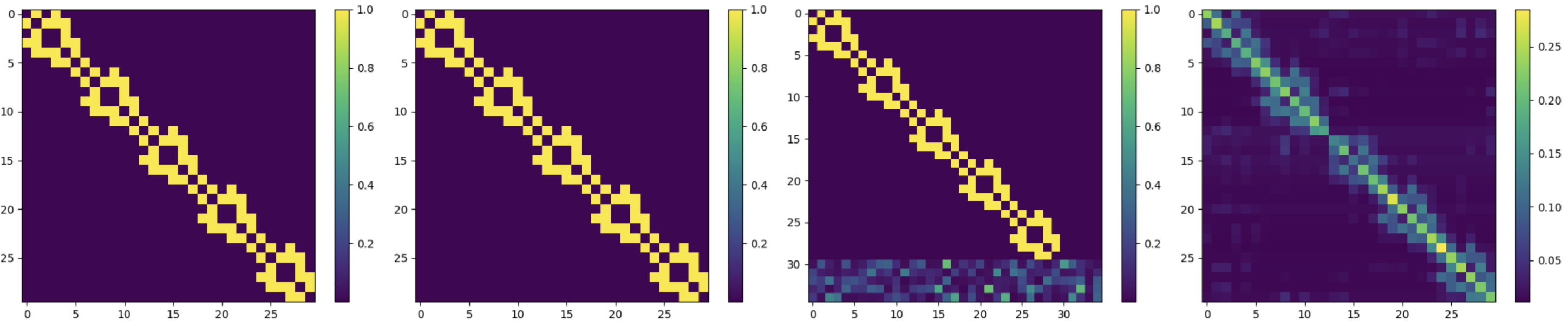}
    \\
    \begin{subfigure}{0.24\textwidth}
        \caption{Ground-truth}% $E(\gx_t)$}
        \label{fig:gt-gpvar-adj}
    \end{subfigure}
    \begin{subfigure}{0.24\textwidth}
        \caption{Id-GSS}
        \label{fig:id-gss-gpvar-adj}
    \end{subfigure}
    \begin{subfigure}{0.24\textwidth}
        \caption{Ext-GSS}
        \label{fig:ext-gss-gpvar-adj}
    \end{subfigure}
    \begin{subfigure}{0.24\textwidth}
        \caption{AGCRN}
        \label{fig:agcrn-gpvar-adj}
    \end{subfigure}
    \caption{Subfigure \ref{fig:gt-gpvar-adj} shows the adjacency matrix of the GPVAR graph. Subfigures~\ref{fig:id-gss-gpvar-adj} and \ref{fig:ext-gss-gpvar-adj} display the probabilities of sampling each edge in the state graph $\gh_t$ learned by Id-GSS and Ext-GSS, respectively.  Subfigure~\ref{fig:agcrn-gpvar-adj} show the edge weight produced by the softmax (see Section~\ref{sec:cast-agcrn}). Node indexing is consistent with Figure~\ref{fig:gpvar}. Nodes with indices $\ge 30$ correspond to the additional state nodes of Ext-GSS.}
    \label{fig:learned-state-graphs}
\end{figure}

\begin{figure}
\centering
\begin{subfigure}[b]{0.18\textwidth}
    \centering
    \includegraphics[height=4cm]{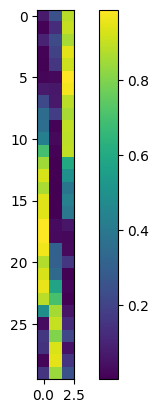}
    \caption{$|\Vh|=3$}
\end{subfigure}
\begin{subfigure}[b]{0.18\textwidth}
    \centering
    \includegraphics[height=4cm]{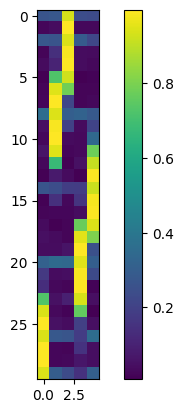}
    \caption{$|\Vh|=5$}
\end{subfigure}
\begin{subfigure}[b]{0.18\textwidth}
    \centering
    \includegraphics[height=4cm]{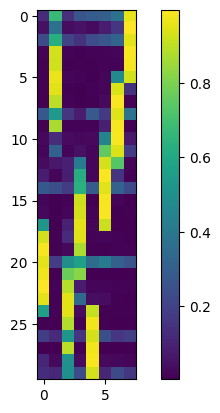}
    \caption{$|\Vh|=8$}
\end{subfigure}
\begin{subfigure}[b]{0.18\textwidth}
    \centering
    \includegraphics[height=4cm]{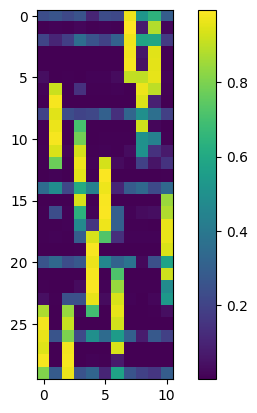}
    \caption{$|\Vh|=11$}
\end{subfigure}
\begin{subfigure}[b]{0.18\textwidth}
    \centering
    \includegraphics[height=4cm]{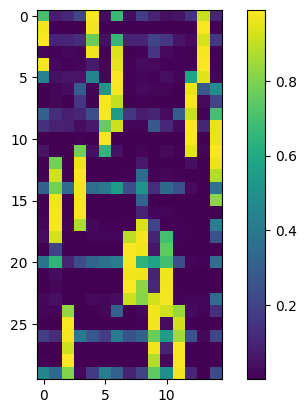}
    \caption{$|\Vh|=15$}
\end{subfigure}
\caption{Probability to assign input nodes (y-axis) to each state node (x-axis), as learned by the Pool-GSS model. 
Each subfigure corresponds to a different number of state nodes: 3, 5, 8, 11, and 15, shown from left to right.}
\label{fig:gpvar-pooling}
\end{figure}

\subsubsection{Results}

Table~\ref{tab:gpvar} reports the forecasting results on the GPVAR dataset.
Baseline models that exploit the ground-truth graph topology $E_{\gx}$ achieve near-optimal MAE, approaching the theoretical optimum of $0.319$ derived from the noise distribution underlying the data-generating process.
The observed discrepancy of approximately $1\%$ can be considered negligible, as further supported by the large $p$-values returned by the AZ-whiteness test, which analyses the prediction residuals to assess model optimality \citep{zambon2022aztest}.
In contrast, the substantially poorer performance of the RNN and fc-RNN baselines highlights the importance of relational structure for accurately solving this prediction task.
Without access to topological information, these models are unable to capture the spatial dependencies governing the dynamics.

Among the proposed GSS configurations, both Id-GSS and Ext-GSS achieve optimal predictive performance as assessed by the AZ-whiteness test despite not receiving the ground-truth graph as input.
Notably, Figure~\ref{fig:learned-state-graphs} shows that these models are able to recover a latent topology that closely matches the ground-truth one used to generate the data.
This result demonstrates that the proposed framework can successfully infer meaningful relational structure directly from time series observations.
By contrast, Pool-GSS and Hub-GSS are unable to retrieve the ground-truth topology, as these models impose a dimensionality bottleneck on the latent state.
Reducing the number of state nodes introduces model approximation errors that lead to suboptimal prediction accuracy.

To further analyze the behavior of Pool-GSS, we train the model with different numbers of state nodes $|\Vh|$.
Figure~\ref{fig:gpvar-pooling} visualizes the learned affiliation matrices, mapping the input nodes in $\Vx$ (y-axis) to the state nodes in $\Vh$ (x-axis).
Since $|\Vh|<|\Vx|$, these mappings are necessarily many-to-one.
The results show that Pool-GSS learns to group together input nodes that are neighbors in the original graph, despite no node relations are provided to the model.
This behavior is consistent with the structure of the data-generating process, in which local information exchange plays a dominant role, while long-range interactions contribute marginally. 

Overall, these results indicate that models instantiated from the proposed GSS framework can be trained effectively and are capable of learning meaningful graph-valued state representations that capture the essential relational structure underlying the observed time series.

\begin{table}
\centering
\begin{tabular}{llccccc}
\toprule
&\multirow{2}{*}{\textbf{Model}} & \multicolumn{2}{c}{\textbf{Relations}} & \multicolumn{3}{c}{\textbf{Test MAE}}\\
& & {Input} & {Learned} & @ 1-6h & @ 1h & @ 6h \\
\midrule
\multirow{5}{0em}{\rotatebox[origin=c]{90}{GSS Baselines}}
& RNN & no & no & 46.933\textsubscript{$\pm$0.508} & 24.953\textsubscript{$\pm$0.243} & 61.027\textsubscript{$\pm$0.712} \\ % 75 K + 0 K + 0 K
& tts-STGNN & yes & no & \rme41.627\textsubscript{$\pm$0.468} & \rgo21.645\textsubscript{$\pm$0.448} & 56.118\textsubscript{$\pm$0.704} \\ % 99 K + 0 K + 0 K
& stt-STGNN & yes & no & 42.685\textsubscript{$\pm$0.364} & \rgo22.267\textsubscript{$\pm$0.397} & 57.020\textsubscript{$\pm$0.518} \\ % 149 K
& t\&s-STGNN & yes & no & 44.167\textsubscript{$\pm$0.364} & 23.166\textsubscript{$\pm$0.437} & 58.906\textsubscript{$\pm$0.452} \\ % 110 K + 0 K + 0 K
& DCRNN & yes & no & 45.257\textsubscript{$\pm$0.575} & 23.359\textsubscript{$\pm$0.377} & 60.267\textsubscript{$\pm$1.139} \\ % 132 K + 0 K + 0 K
\midrule
\multirow{5}{0em}{\rotatebox[origin=c]{90}{GSS Models}}
& Id-GSS & no & yes & 43.375\textsubscript{$\pm$0.341} & 24.756\textsubscript{$\pm$0.482} & 55.984\textsubscript{$\pm$0.635} \\ % 136 K + 0 K + 237 K
& Ext-GSS & no & yes & 43.333\textsubscript{$\pm$0.252} & 24.594\textsubscript{$\pm$0.243} & 55.882\textsubscript{$\pm$0.531} \\ % 129 K + 7 K + 252 K
& Pool-GSS & no & yes & \rme41.288\textsubscript{$\pm$0.206} & 23.632\textsubscript{$\pm$0.216} & \rme54.458\textsubscript{$\pm$0.409} \\ % 169 K + 7 K + .2 K
& Hub-GSS & yes & yes & \rgo40.014\textsubscript{$\pm$0.109} & \rme22.705\textsubscript{$\pm$0.540} & \rgo52.758\textsubscript{$\pm$0.258} \\ % 122 K + 7 K + .0 K
& AGCRN & no & yes & 42.744\textsubscript{$\pm$0.752} & 25.737\textsubscript{$\pm$0.410} & \rme54.956\textsubscript{$\pm$1.079} \\ % 988 K + 0 K + 4 K
\midrule
\multirow{4}{0em}{\rotatebox[origin=c]{90}{Not GSS}}
& GWNet & yes & yes & 39.482\textsubscript{$\pm$0.208} & 22.027\textsubscript{$\pm$0.392} & 52.289\textsubscript{$\pm$0.398} \\ % 292 K + 0 K + 9 K
& GWNet-ig & yes & no & 42.053\textsubscript{$\pm$0.186} & 22.021\textsubscript{$\pm$0.286} & 56.654\textsubscript{$\pm$0.536} \\
& GWNet-gsl & no & yes & 39.927\textsubscript{$\pm$0.357} & 22.337\textsubscript{$\pm$0.328} & 52.662\textsubscript{$\pm$0.402} \\
& tts-Transf & no & no & 40.289\textsubscript{$\pm$0.285} & 23.715\textsubscript{$\pm$0.450} & 52.365\textsubscript{$\pm$0.588} \\ % 172 K
\bottomrule
\end{tabular}
\caption{Results on 6-step ahead forecasting on EngRAD. Reported values are averages of 5 independent runs with standard deviation reported subscripted. The best GSS model performance for each column is highlighted in green and close to best (<5\% from the best one) in yellow.}
\label{tab:engrad}
\end{table}

\begin{figure}
    \centering
    \begin{subfigure}{0.34\textwidth}
        \includegraphics[width=1\textwidth]{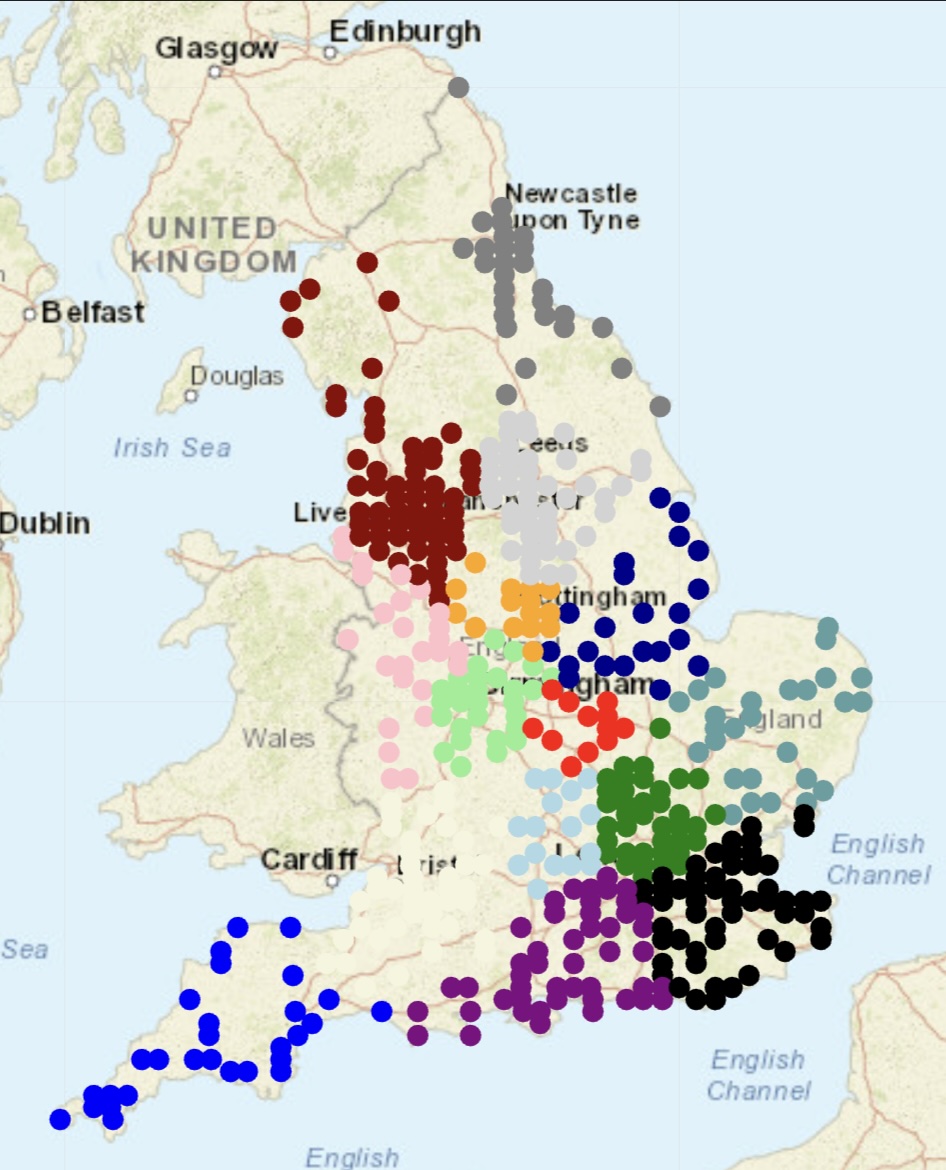}
        \caption{Pool-GSS}
        \label{fig:leading-node-assignment-pool}
    \end{subfigure}
    \begin{subfigure}{0.34\textwidth}
        \includegraphics[width=1\textwidth]{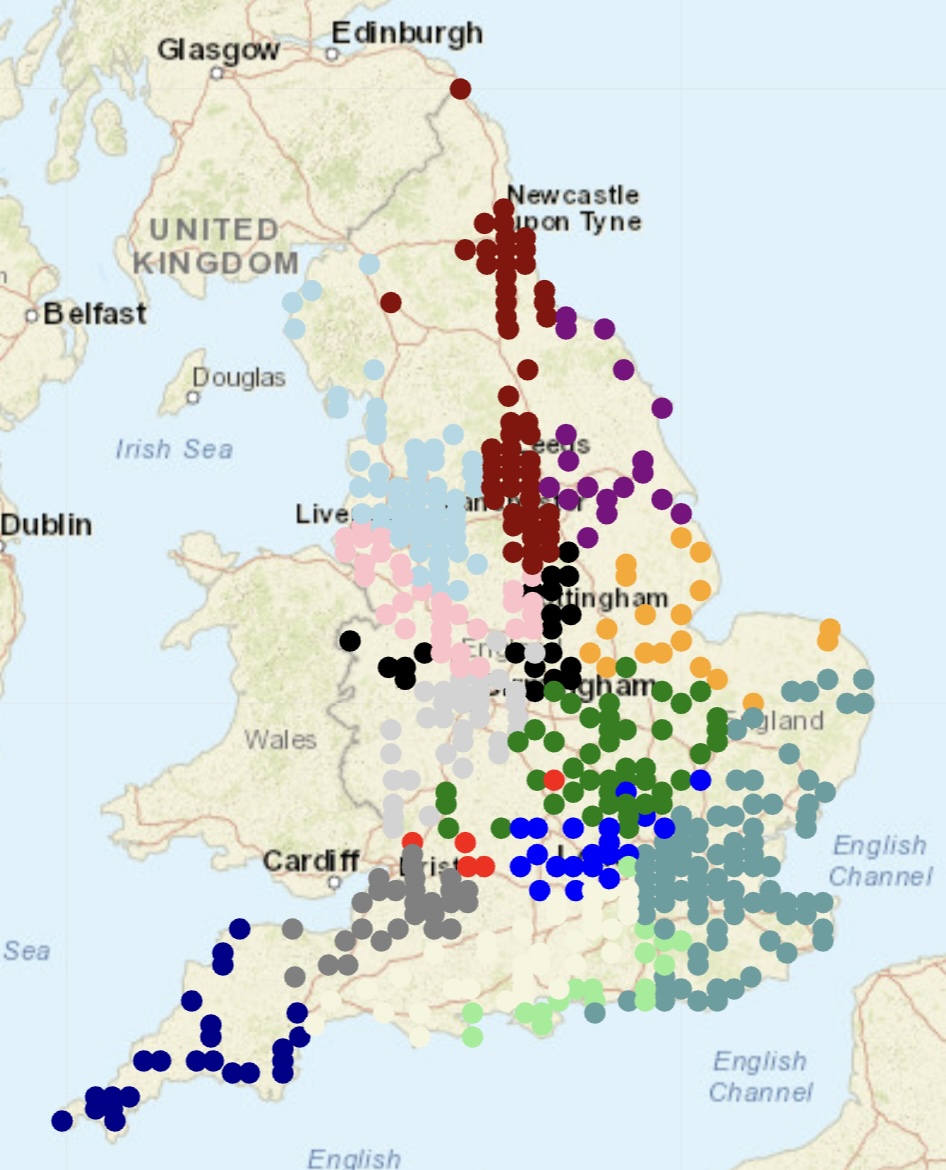}
        \caption{Hub-GSS}
        \label{fig:leading-node-assignment-hub}
    \end{subfigure}
    \caption{Visualization of the learned correspondence between input and state nodes. The figure shows one learned instance of Pool-GSS \eqref{fig:leading-node-assignment-pool} and one of Hub-GSS \eqref{fig:leading-node-assignment-hub}, with the number of state nodes $\Vh$ set to 15; the specific model instances are randomly selected among the five runs. 
    Points represent the input nodes of the EngRAD dataset, while colors denote the state nodes. The color assigned to each input node corresponds to the state node with the highest score in the learned affiliation matrix $\vect S$, \ie, the node associated with the maximum value along each column of $\vect S$. There is no color correspondence between the two figures.}
    \label{fig:leading-node-assignment}
\end{figure}

\subsection{Solar Radiation Forecasting}

\subsubsection{Problem Setup}
In this experiment, we study a multi-step forecasting task related to an energy production application. 
We use the EngRAD dataset introduced in \citep{marisca2024graph}, which contains five weather variables related to solar radiation sampled every hour at 487 locations across England over three years.
We consider the global horizontal irradiation as the target variable to be predicted six hours ahead at each location. All five variables from the preceding day, additional encodings of the hour of the day and hour of the year, and a mask of the missing observations are considered as input signals. 

There is no obvious graph to consider in graph-based processing for this task. This is indeed one main advantage of GSS models, where relational information can be learned from data. 
For those models requiring an input graph to operate, we connect nodes according to their spatial proximity \citep{marisca2024graph}, as commonly done in the related literature. In particular, we keep the 8 nearest neighbors of each node with a distance smaller than 76 km.
Models are trained to minimize the MAE on all the six lead times. The first two years of data are used for training, with twelve weeks of data across the second year reserved for validation, and the last year is used for testing. 

\subsubsection{Results}

Table~\ref{tab:engrad} reports the forecasting performance on the EngRAD dataset, measured in terms of MAE aggregated over the entire 6-hour horizon (MAE @ 1--6h), as well as at the first and last prediction steps (MAE @ 1h and MAE @ 6h).

The first observation is that overall the lowest prediction errors are obtained with models that account for spatial dependencies. The RNN, instead, operates on independent time series and shows the weakest performance. 
In particular, we see that good performance is achieved by models using the input graph (\eg, tts-STGNN), only learned relations (see Pool-GSS and tts-Transf), and combinations of the two (such as Hub-GSS and GWNet).
This trend suggests that there are meaningful relations that are informative to solve this problem and, therefore, the appropriateness of this setup to study GSS models.

Analyzing in more detail the GSS models and baselines, the best short-term predictions (see MAE @ 1h) are provided by tts-STGNN and t\&s-STGNN variants, relying on the input graph based on spatial proximity. This result appears consistent with the local nature of the dynamics for small time increments.
Conversely, Pool-GSS and AGCRN models rely solely on learned relations and rank among the best on longer-term predictions. Notably, Pool-GSS maintains strong performance despite its architectural bottleneck. 
Consistent with these observations, the Hub-GSS model, which extends Pool-GSS by accounting for the input graph at the level of the encoder (see Table~\ref{tab:models}), improves over Pool-GSS both on the long and short terms, achieving the best overall predictions within the GSS family. The principal correspondence between input nodes and state nodes learned   by Pool-GSS and Hub-GSS are depicted in Figure~\ref{fig:leading-node-assignment}, evidencing the extraction of local structures; the figure supports our claim that the relational information learned or inferred by GSS models can be meaningful and provide valuable insights for the user.
A similar behavior on the prediction performance can be seen by comparing GWNet with its variants without learned relational information (GWNet-ig, with only the input graph) and with only the learned relational information (GWNet-gsl). 
A possible motivation for these results is twofold. On the one hand, although proximity is an effective inductive bias for very short-term predictions, it becomes less informative at longer horizons, where global conditions become more relevant. Loss-driven relational structures learned by the GSS models can compensate purely local connection. On the other hand, adopting MAE @ 1--6h as the training objective may implicitly place greater emphasis on optimizing performance at longer lead times (approximately 3--6h ahead), which tend to constitute more homogeneous prediction tasks than the very short-term horizons (1h or 2h ahead). 

Taken together, the results of this section show that the proposed framework provides a practical and flexible approach to spatio-temporal forecasting in settings where relational structure is unknown or not fully available.
The different GSS configurations adapt to the task by learning latent interactions, aggregating information at appropriate spatial scales, where needed.
The general GSS formulation allows tailoring the models to better serve the given application setting, implementing specific architectural biases, when these are known. At the same time, it allows the model to explicitly learn latent relations that help the prediction task at hand.
Importantly, these capabilities emerge without sacrificing predictive accuracy, as the resulting models remain competitive with different architectures, from convolutional models to transformers.
This balance between modeling flexibility, interpretability of the learned structures, and empirical performance highlights the practical value of the proposed graph state-space formulation.

\section{Conclusions}

This paper introduced a general probabilistic framework for graph state-space modeling, aimed at unifying state-space representations and graph-based learning for spatio-temporal data. By representing inputs, latent states, and outputs as graphs whose topology and node sets may differ and evolve, the proposed formulation extends classical state-space models to settings in which relational structure is unknown, latent, or task-dependent.
This design supports end-to-end learning of both node-level dynamics and discrete relational structures, while remaining flexible with respect to architectural choices and implementation details. Several existing methods from the literature were reinterpreted as special cases of the proposed formulation, highlighting its generality and clarifying the relationships among a broad class of spatio-temporal models.

Experimental results on synthetic and real-world datasets demonstrated that models derived from the proposed framework can be trained effectively despite the presence of latent and discrete variables. On controlled synthetic data, GSS models were shown to recover meaningful relational structure and achieve optimal or near-optimal predictive performance.
On a larger-scale solar radiation forecasting task, the same framework supported multiple architectural instantiations that adapt to the problem requirements, yielding performance competitive with existing spatio-temporal graph neural networks and transformer-based approaches.

The proposed graph state-space formulation offers a principled and versatile approach to modeling spatio-temporal processes, bridging probabilistic state-space modeling and relational learning. Future work may explore extensions to graph generative settings, deepen the study of the learnability of dynamic relations, and systematically integrate domain-specific constraints, further broadening the applicability of the framework to complex real-world systems. 
Finally, the development of Kalman filtering and smoothing on GSS models constitutes a promising research direction \citep{alippi2023graph}, with significant potential in modeling, state estimation and control of complex systems from domains like meteoreology, energy managament, and transportations.

\subsection*{Acknowledgements}
This work was partly supported by the Swiss National Science Foundation projects no.\ 204061~(High-Order Relations and Dynamics in Graph Neural Networks) and no.\ 225351~(Relational Deep Learning for Reliable Time Series Forecasting at Scale). The authors thank Lorenzo Livi for his valuable insights and discussion on this work.

\appendix

\section{Model Hyperparameters} \label{app:hparams}

For GPVAR, the GSS models have a linear encoder, a 2-layer state transition, and an MLP with 1 hidden layer as the readout; for the stt-STGNN, 2 encoding layers and 1 state-transition layer are implemented. Unless specified otherwise, models use the ELU as activation function. Pool-GSS relies on 5 state nodes, and Ext-GSS extends the state by 5 nodes. The message-passing layers mean-aggregates node features and apply the Tanh activation on top of a linear feature transformation. The hidden size is set to 32. The learnable node embeddings have size 8 and are passed to both the encoder and the readout.
The RNN and fc-RNN models have a 2-layer encoder, 2 recurrent layers, a 2-hidden-layer readout, and 64 hidden neurons. For EngRAD, the hidden size is set to 64 (128 for the RNN) and the embedding size to 32. Pool-GSS has 15 state nodes, and Ext-GSS adds 15 nodes. Encoders with message passing are 2-layer, the state transition is 1-layer, and the message-passing operator is Diffusion Convolution \citep{li2018diffusion}. The tts-STGNN has 4 message-passing layers. 
The spatiotemporal transformer tts-Transf applies, node-wise, a 3-layer transformer with causal attention along the temporal dimension; spatial processing is then performed across nodes via a second 2-layer transformer. The transformer interleaves 2-head attention layers with a hidden size of 64 and a 2-layer MLP with a hidden-size of 128. The model relies on layer normalization and 8-dimensional learnable node embeddings passed as additional input to the model.
All models are trained to minimize the MAE for a maximum of $200$ epochs with the Adam optimizer, early stopping monitoring the validation MAE, and halving the learning rate after 10-epoch plateaus in the loss.
Initial learning rates are $0.01$ for GPVAR and $0.001$ for EngRAD.

\section{Hardware and Software}

The code for the empirical evaluation of the proposed method has been developed in Python relying on the following open-source libraries: PyTorch~\citep{paske2019pytorch}, PyTorch Geometric~\citep{fey2019fast}, Torch Spatiotemporal~\citep{Cini_Torch_Spatiotemporal_2022}, PyTorch Lightning~\citep{Falcon_PyTorch_Lightning_2019} and NumPy~\citep{harris2020array}. The experiments were run on machines equipped with Intel(R) Xeon(R) CPU, 192 GB RAM, and NVIDIA L4 GPU. The code for reproducing all the experiments is released upon publication.

\bibliographystyle{plainnat}
\bibliography{pan}

\end{document}